\documentclass[letterpaper, 10 pt, conference]{ieeeconf}  % Comment this line out if you need a4paper

\IEEEoverridecommandlockouts                              % This command is only needed if 
\usepackage{graphicx}
\usepackage{subfig}
\graphicspath{ {figures/} }
\usepackage{amsmath} % assumes amsmath package installed
\usepackage{amssymb}  % assumes amsmath package installed
\usepackage{amsfonts}
\let\labelindent\relax
\usepackage{enumitem}
\usepackage{booktabs}

\title{\LARGE \bf
ARiADNE: A Reinforcement learning approach using Attention-based Deep Networks for Exploration
}

\author{Yuhong Cao$^{1}$, Tianxiang Hou$^{1}$, Yizhuo Wang$^{1}$, Xian Yi$^{1}$, Guillaume Sartoretti$^{1}$$^{\dagger}$
\thanks{$\dagger$ Corresponding author, to whom correspondence should be addressed.}
\thanks{$^{1}$ Authors are with the Department of Mechanical Engineering, College of Design and Engineering, National University of Singapore.
{\tt\small caoyuhong@u.nus.edu, htx24@foxmail.com, \{wy98,yxian11\}@u.nus.edu, mpegas@nus.edu.sg}}}

\begin{document}

\maketitle
\thispagestyle{empty}
\pagestyle{empty}

%%%%%%%%%%%%%%%%%%%%%%%%%%%%%%%%%%%%%%%%%%%%%%%%%%%%%%%%%%%%%%%%%%%%%%%%%%%%%%%%

\begin{abstract}

In autonomous robot exploration tasks, a mobile robot needs to actively explore and map an unknown environment as fast as possible. Since the environment is being revealed during exploration, the robot needs to frequently re-plan its path online, as new information is acquired by onboard sensors and used to update its partial map. While state-of-the-art exploration planners are frontier- and sampling-based, encouraged by the recent development in deep reinforcement learning (DRL), we propose ARiADNE, an attention-based neural approach to obtain real-time, non-myopic path planning for autonomous exploration. ARiADNE is able to learn dependencies at multiple spatial scales between areas of the agent's partial map, and implicitly predict potential gains associated with exploring those areas. This allows the agent to sequence movement actions that balance the natural trade-off between exploitation/refinement of the map in known areas and exploration of new areas. We experimentally demonstrate that our method outperforms both learning and non-learning state-of-the-art baselines in terms of average trajectory length to complete exploration in hundreds of simplified 2D indoor scenarios. We further validate our approach in high-fidelity Robot Operating System (ROS) simulations, where we consider a real sensor model and a realistic low-level motion controller, toward deployment on real robots.

\end{abstract}

%%%%%%%%%%%%%%%%%%%%%%%%%%%%%%%%%%%%%%%%%%%%%%%%%%%%%%%%%%%%%%%%%%%%%%%%%%%%%%%%
%%%%%%%%%%%%%%%%%%%%%%%%%%%%%%%%%%%%%%%%%%%%%%%%%%%%%%%%%%%%%%%%%%%%%%%%%%%%%%%%

\section{INTRODUCTION}

Autonomous robot exploration (ARE) refers to the task, in which a robot needs to autonomously explore and map an unknown environment as efficiently and quickly as possible. The robot is usually equipped with a sensor (e.g., LiDAR or camera) to obtain measurements of its surroundings and build/update a \textit{partial} map of the environment. In practice, the (high-dimensional) collected sensor data (e.g., point cloud) is usually converted into a (simplified) occupancy grid map or Octomap~\cite{hornung13auro} that can be used for further planning~\cite{placed2022survey,bircher2016receding,cao2021tare}.
% During exploration, the robot keeps updating its partial map until the environment is fully explored.
Such a task is also known as active SLAM~\cite{placed2022survey}.  
Although a robot can always slowly but accurately construct a map by carefully methodically covering the entire environment, the objective of ARE is to plan the shortest path to complete exploration, where small noises/errors in the final map are tolerable. In consequence, the main challenge of ARE is to plan a \textit{non-myopic} path that balances the trade-off between exploiting surroundings (i.e., refining the map in already-explored areas) and exploring new (usually, further away) areas, most importantly with only partial knowledge of the environment. Such an exploration path is usually planned incrementally online, as the partial map is updated using new measurements along the way.

For example, conventional frontier-based methods~\cite{bircher2016receding,yamauchi1997frontier,gonzalez2002navigation,selin2019efficient} generate multiple candidate paths, each covering a \textit{frontier} (i.e., the boundary between explored free area and unexplored area), and greedily select the path with maximum \textit{gain}, usually defined as a combination of \textit{utility} (i.e., number of observable frontiers along the path) and \textit{cost} (i.e., path length). However, an essential problem of these methods is that such myopic frontier selection does not guarantee optimality in the long term. A more recent approach~\cite{cao2021tare} reasons about the whole path to cover all current frontiers, thus guaranteeing (near-)optimal paths in the current partial map. However, since the environment is only partially known, a previous optimal path often quickly becomes sub-optimal as more of the environment is revealed, or even worse, results in redundant movements  (e.g., missing an unexplored shortcut between two rooms which were previously not known to be connected). Based on experiences from conventional exploration planners, we note that \textit{non-myopicity} comes at two different levels in autonomous exploration. The first level, \textit{spatial non-myopicity}, requires the planner to reason about the current partial map to balance the exploration-exploitation trade-off, while \textit{temporal non-myopicity} requires the planner to estimate the future influence of current decisions (e.g., predict the changes in the partial map that may stem from given path planning decisions).

\begin{figure}[t]
    \vspace{0.2cm}
    \centering
    \includegraphics[width=0.4\textwidth]{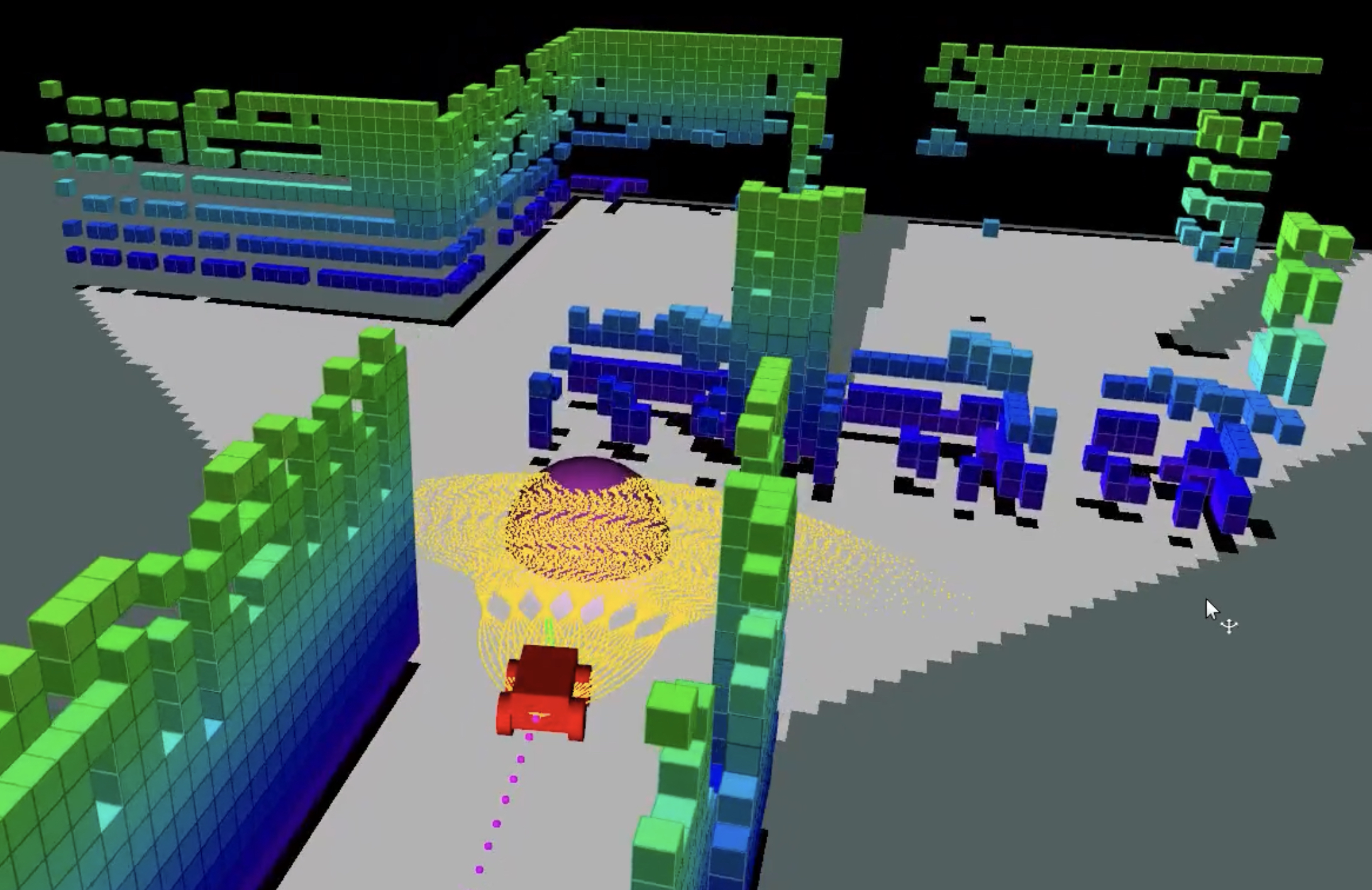}
    \vspace{-0.2cm}
    \caption{\textbf{Illustration of autonomous robot exploration.} A wheeled robot is building a 3D Octomap using an onboard LiDAR. The purple ball indicates the next viewpoint output by our planner, which is tracked by the robot using a low-level motion controller (yellow primitives).}
    \label{fig:example}
    \vspace{-0.45cm}
\end{figure}

To achieve non-myopicity at these two levels, we propose a deep reinforcement learning (DRL) based approach for ARE, named ARiADNE, which relies on two attention-based neural networks. These networks allow the agent to reason about dependencies of different areas in the partial map at different spatial scales, thus allowing the agent to sequence spatially non-myopic decisions efficiently without the need to optimize a long path. Furthermore, our critic network implicitly provides the robot with the ability to estimate potential areas that might be found by learning the state-action value, which furthermore helps make decisions beneficial to the long-term efficiency, thus addressing temporally non-myopicity. Specifically, we first formulate autonomous exploration as a sequential decision-making problem on a collision-free graph that covers the known traversable area, where one of the nodes is the robot's current position. We then use our attention-based neural network to select one neighboring node of the current robot position as the next viewpoint for the robot. In this work, we focus on training and testing our approach in indoor environments based on 2D occupancy grid maps, ranging from simple scenarios (single room) to relatively complex ones (multiple rooms with complicated corridors). There, we experimentally demonstrate that our approach outperforms state-of-the-art conventional methods on average. We also validate our approach in high-fidelity ROS (Robot Operating System) simulations, where we consider a real sensor model and a motion controller, showing the generalizability of our model to realistic environments.

%%%%%%%%%%%%%%%%%%%%%%%%%%%%%%%%%%%%%%%%%%%%%%%%%%%%%%%%%%%%%%%%%%%%%%%%%%%%%%%%
%%%%%%%%%%%%%%%%%%%%%%%%%%%%%%%%%%%%%%%%%%%%%%%%%%%%%%%%%%%%%%%%%%%%%%%%%%%%%%%%

\section{RELATED WORK}

\begin{description}[leftmargin=*]

\item[Frontier-based vs sampling-based approaches]
The first frontier-based method was proposed by Yamauchi~\cite{yamauchi1997frontier}, in which the robot is constantly driven towards the nearest frontier. In more advanced frontier-based methods, selections of which frontier to visit are evaluated by a gain function, which considers the effect of both utility and cost~\cite{gonzalez2002navigation,holz2010evaluating, kulich2011distance}. On the other hand, the past few years have seen a number of sampling-based methods be developed, based on Rapidly-exploring Random Trees (RRT)~\cite{bircher2016receding}, Rapidly-exploring Random Graphs~\cite{dang2020graph}, and Probabilistic Random Maps (PRM)~\cite{xu2021autonomous}. Sampling-based methods only need to compute the gain of sampled paths, which avoids the complexity of identifying and evaluating all frontiers. Recent works demonstrated that frontier-based methods are more suitable when frontiers are \textit{sparse} (e.g., 2D indoor scenarios), while sampling-based methods perform better with \textit{dense} frontiers (e.g., 3D outdoor scenarios)~\cite{selin2019efficient,xu2021autonomous}. Intuitively, Frontier-based methods become inefficient when there are too many frontiers to evaluate, while sampling-based methods underperform when informative paths are hard to sample.

\item[Planning for long-term objectives]
Both frontier-based and sampling-based methods mostly rely on greedy strategies to plan short-term paths. Since the robot only has access to a partial map of the environment, such paths inevitably lead to myopic performance in the long term. Recently, Cao et al.~\cite{cao2021tare} proposed TARE to optimize the full exploration path and mitigate myopicity. Utilizing the full knowledge of the current partial map, TARE was shown to significantly outperform state-of-the-art sampling-based planners in large-scale 3D scenarios. Similar methods were also considered to approach the \textit{informative path planning}~\cite{arora2017randomized} problem, which considers information gathering usually in obstacle-free environments (in fact, works on autonomous exploration and informative path planning mutually promote each other, e.g.,~\cite{cao2021tare} and~\cite{arora2017randomized},~\cite{hollinger2014sampling} and~\cite{bircher2016receding}).

\item[Learning-based exploration]

Niroui et al.~\cite{niroui2019deep} first combined frontier-based method with deep reinforcement learning to adaptively tune the parameter of the gain function for frontier selections and improve performance. Schmid et al.~\cite{schmid2022fast} proposed to learn the underlying gain distribution based on the partial map by supervised learning, thus helping sampling-based methods more efficiently find next-best-views and reduce computation. While the above works focus on improving conventional methods using machine learning, some other works~\cite{zhu2018deep,li2019deep,chen2019self,chen2020autonomous} directly applied deep reinforcement learning to select a viewpoint to visit, often relying on convolutional neural networks (CNNs). However, although~\cite{zhu2018deep,chen2019self} argue that DRL-based methods naturally optimize long-term objectives, it seems that~\cite{zhu2018deep,li2019deep,chen2019self,chen2020autonomous} are only able to reach performance slightly better than the nearest-frontier method so far.

\end{description}

%%%%%%%%%%%%%%%%%%%%%%%%%%%%%%%%%%%%%%%%%%%%%%%%%%%%%%%%%%%%%%%%%%%%%%%%%%%%%%%%
%%%%%%%%%%%%%%%%%%%%%%%%%%%%%%%%%%%%%%%%%%%%%%%%%%%%%%%%%%%%%%%%%%%%%%%%%%%%%%%%

\section{PROBLEM FORMULATION}

We consider a bounded and unknown environment $\mathcal{E}$ represented by a $x \times y$ 2D occupancy grid map, whose partial (occupancy grid) map is denoted as $P$. The partial map consists of unknown area $P_u$ (i.e., unexplored area) and known area $P_k$ (i.e., explored area), such that $P_u\cup P_k=P$. The known area $P_k$ is further classified into free area $P_f$ (i.e., traversable area for the robot) and occupied area $P_o$ (i.e., obstacles) such that $P_f \cup P_o=P_k$. At the beginning of exploration, the environment is fully unknown so the partial map $P=P_u$. Then, during exploration, the unknown area in the sensor range $d_s$ (the sensor we use is a 360-degree LiDAR) is classified into either free area or occupied area according to sensor measurements.
The objective of autonomous exploration is to find the shortest collision-free robot trajectory $\psi^*$ to complete exploration:
\vspace{-0.15cm}
\begin{equation}
    \psi^*= \mathop{\rm {argmin}} \limits_{\psi \in \Psi}{\rm C}(\psi),\ {\rm {s.t.}}\ P_k=P_g,
    \label{eq:objective}
\vspace{-0.15cm}
\end{equation}
where $\rm C: \psi \xrightarrow{} \mathbb{R}^+$ maps a trajectory to its length and $P_g$ denotes the ground truth of the environment. Although the ground truth is not accessible in real-world deployments, it is known and can be utilized to evaluate the performance of planners in testing. In practice, most works consider the closure of occupied areas as $P_k=P_g$~\cite{yamauchi1997frontier,cao2021tare,chen2019self,chen2020autonomous}.

%%%%%%%%%%%%%%%%%%%%%%%%%%%%%%%%%%%%%%%%%%%%%%%%%%%%%%%%%%%%%%%%%%%%%%%%%%%%%%%%
%%%%%%%%%%%%%%%%%%%%%%%%%%%%%%%%%%%%%%%%%%%%%%%%%%%%%%%%%%%%%%%%%%%%%%%%%%%%%%%%

\section{METHOD}

In this section, we cast ARE as an RL problem, and introduce our attention-based policy and critic neural networks as well as details of our training.

%%%%%%%%%%%%%%%%%%%%%%%%%%%%%%%%%%%%%%%%%%%%%%%%%%%%%%%%%%%%%%%%%%%%%%%%%%%%%%%%

\subsection{Exploration as an RL Problem}

\begin{description}[leftmargin=*]

\item[Sequential Decision-making Problem]
Since the free area is updated based on the robot's movements, online planning for ARE is a sequential decision-making problem in nature. Following our previous work~\cite{cao2022catnipp} for informative path planning, we consider the robot trajectory $\psi$ as a sequence of viewpoints $\psi=(\psi_0, \psi_1, ...), \psi_i\in P_f$. At each decision step $t$, we first uniformly distribute candidate viewpoints $V_t=\{v_0, v_1, ...\}$, $\forall \ v_i=(x_i, y_i) \in P_f$ in the current free area $P_f$, similar to~\cite{cao2021tare}. Then, to find collision-free paths between viewpoints, we connect each viewpoint with its $k$ nearest neighbors through a straight line and remove edges that collide with occupied or unknown areas. In doing so, we build a collision-free graph $G_t=(V_t, E_t)$, with $V_t$ a set of uniformly distributed nodes (i.e., viewpoints) over the free area, and $E_t$ a set of traversable edges. We finally let the robot select one neighboring node of its current position $\psi_t$ as the next viewpoint. Since the decision will be taken upon arriving at the last selected viewpoint, the trajectory is a sequence of waypoints such that $\psi_i \in V$.

\item[Observation]
The observation of the agent is $o_t=(G'_t, \psi_t)$, where $G'_t=(V'_t, E_t)$ is the \textit{augmented graph} based on the current collision-free graph $G_t$, while $\psi_t$ is the robot current position. Note that $G'_t$ shares the same edge set $E_t$ as $G_t$. In addition to the node coordinates (i.e., $v_i=(x_i, y_i)$), The properties of each node $v'_i$ in the augmented graph further include a binary signal $b_i$, which indicates if the node has been visited by the agent already, and the associated utility $u_i$, such that $v'_i=(x_i, y_i, u_i, b_i)$.
We experimentally found that the binary signal helps improve the learning by allowing the robot to be aware of its previous movements.
The utility $u_i$ represents the number of observable frontiers at node $v_i$~\cite{cao2021tare}. We consider observable frontiers as frontiers within \textit{light of sight} of the node (i.e., lines between the node and observable frontiers are collision-free and their length is smaller than the sensor range).
The utility $u_i$ at node $v_i$ is computed as $u_i=|F_{o,i}|, \forall f_j \in F_{o,i}, ||f_j-v_i||\leq d_s, L(v_i, f_j) \cap (P-P_f) = \emptyset$,
where $F_{o,i}$ denotes the observable frontiers set at node $v_i$, $d_s$ denotes the sensor range and $L(v_i, f_j)$ the line between node $v_i$ and frontier $f_j$.
In practice, we scale the node coordinates and utility to $[0,1]$ before feeding the observation into the neural network.

\item[Action] At each decision step $t$, given the agent's observation $o_t$, our attention-based neural network outputs a stochastic policy to select a node out of all neighboring nodes as the next viewpoint to visit. The policy is denoted as $\pi_{\theta}(a_t|o_t) = \pi_{\theta}(\psi_{t+1}=v_i,(\psi_t, v_i)\in E_t \mid o_t)$, where $\theta$ represents the set of weights of the neural network. The robot moves to the next viewpoint in a straight line, and updates its partial map based on data collected along the way.

\item[Reward]
To encourage efficient exploration, after taking each movement action $a_t$, the robot receives a reward composed of three parts. The first part $r_o=|F_{o, \psi_{t+1}}|$ is the number of observed frontiers at the new viewpoint. The second part $r_c=-\rm{C}(\psi_t, \psi_{t+1})$ is a punishment on the distance between the previous and new viewpoints. A fixed finishing reward $r_f=\left\{\begin{array}{ll}20, & P_k=P_g\\ 0, &\rm{otherwise,} \end{array}\right.$ is given at the end of the episode, if and only if the exploration task was completed. The total reward reads: $r_t(o_t, a_t)=a\cdot r_o + b\cdot r_c + r_f$, where $a$ and $b$ are scaling parameters (in practice $a=1/50, b=1/64$).

\end{description}

\begin{figure}[tb]
    \vspace{0.2cm}
    \centering
    \includegraphics[width=0.35\textwidth]{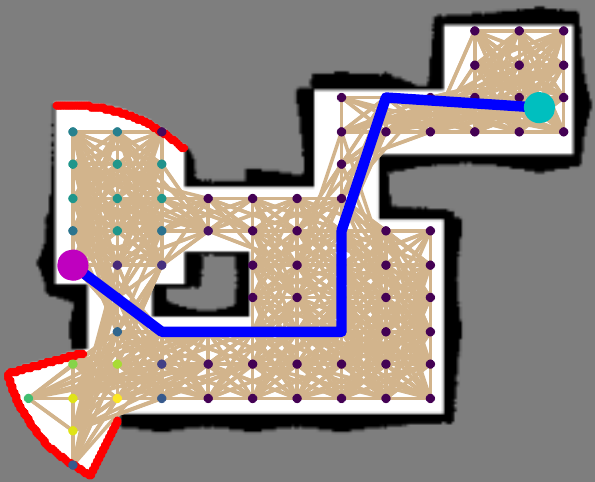}
    % \vspace{-0.1cm}
    \caption{\textbf{Example decision step in the middle of an exploration task in our approach}, showing the unknown area (grey cells), free area (white cells), occupied area (black cells), frontiers (red cells), executed trajectory (blue line), graph edges (tan lines), candidate viewpoints (small dots, whose color represents their utility), robot current position (purple disk), and robot starting position (light blue disk).}
    \label{fig:formulation}
    \vspace{-0.45cm}
\end{figure}

\begin{figure*}[tb]
    \centering
    \includegraphics{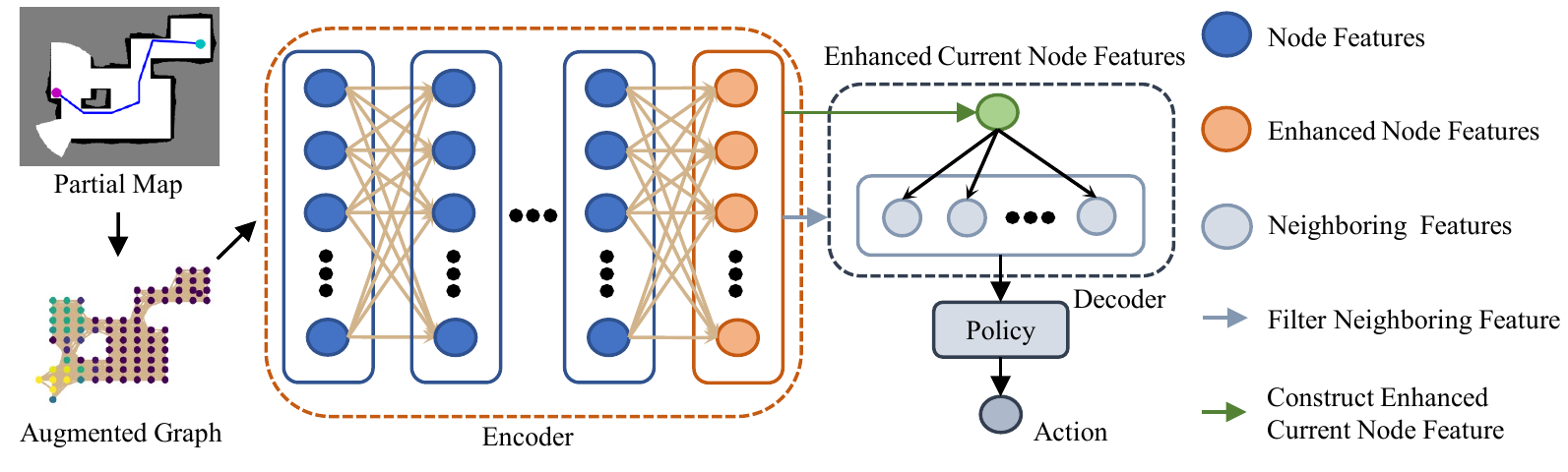}
    \vspace{-0.3cm}
    \caption{\textbf{Attention-based policy network.} Note that neighboring relationships in the augmented graph (tan) are also used as the mask~\cite{vaswani2017attention} in attention layers in the encoder.}
    \label{fig:model}
    \vspace{-0.45cm}
\end{figure*}

%%%%%%%%%%%%%%%%%%%%%%%%%%%%%%%%%%%%%%%%%%%%%%%%%%%%%%%%%%%%%%%%%%%%%%%%%%%%%%%%

\subsection{Policy Network}

The policy $\psi_\theta$ is output by our attention-based neural network, which is composed of an encoder and a decoder (shown in Fig.~\ref{fig:model}).
We first rely on the encoder to extract salient features from the current partial map, specifically by learning dependencies between nodes in the associated augmented graph $G'$.
Based on these features as well as the current robot position, the decoder then outputs the policy over neighboring nodes, which can be used to decide which one to visit next.
Note that, while policy-based RL agents often have a fixed action space, our decoder is inspired by the Pointer Network~\cite{vinyals2015pointer} to allow the action space to depend on the number of neighboring nodes input in the network.
This allows our network to naturally adapt to our collision-free graph, where nodes have arbitrary numbers of neighbors.

\begin{description}[leftmargin=*]

\item[Attention Layer]
We use the attention layer~\cite{vaswani2017attention} as the fundamental building block in our model. The input of such an attention layer is composed of a query vector $h^q$ and a key-and-value vector $h^{k,v}$. The output of this layer, $h'_i$, is the weighted sum of the value vector, where weights depend on the similarity between key and query:

\begin{equation}
\begin{aligned}
& q_{i}=W^Qh^{q}_{i}, \ k_{i}=W^Kh^{k,v}_{i}, \ v_{i}=W^Vh^{k,v}_{i}, \\
& u_{ij}=\frac{q_{i}^T\cdot k_{j}}{\sqrt{d}}, \  w_{ij}=\frac{e^{u_{ij}}}{\sum_{j=1}^{n}e^{u_{ij}}}, \   h'_{i}=\sum_{j=1}^{n}w_{ij}v_{j},
\end{aligned}
\label{eq:attention}
\end{equation}

where $W^Q, W^K, W^V \in \mathbb{R}^{d\times d}$ are all learnable matrices. Updated features are then passed through a feed-forward sublayer, following~\cite{vaswani2017attention}.

\item[Encoder]
In the encoder, we first linearly embed the \textit{node inputs} $V'$ into $d$-dimensional \textit{node features} $h^n$, where $h^n_i=W^lv'_i+b^l$. We then calculate an edge mask $M$ where $m_{ij}=\left\{\begin{array}{l}0,\ (v_i, v_j) \in E_t\\1 ,\ (v_i, v_j) \notin E_t \end{array}\right.$. The node features are then passed to multiple (6 in practice) stacked attention layers, where $h^q=h^{k,v}=h^n$, each attention layer taking the output of the previous one as input. An edge mask is applied to allow each node access to its neighboring node features only, by setting $w_{ij}=0, \ \forall (i,j),\ m_{ij}=1$. Despite attention being restricted to neighboring nodes in each layer, nodes can still obtain non-neighboring node features by aggregating node features multiple times through this stacked attention structure. We empirically found that such structure is more suitable than graph transformers~\cite{dwivedi2020generalization} (like in our previous work~\cite{cao2022catnipp}) to learn path finding in maps with cluttered obstacles. We term the output of the encoder, $\hat{h}^{e}$, the \textit{enhanced node features}, since each of these updated node features $\hat{h}^n_i$ contains the dependencies of $v'_i$ with other nodes.

\item[Decoder]
We use the decoder to output a policy based on enhanced node features $\hat{h}^{e}$ and the current robot position $\psi_t$. Denoting the current robot position as node $v_c=\psi_t$, we first select the \textit{current node features} $h^c$ and \textit{neighboring features} $h^{nb}$, $\forall \hat{h}^{nb}_i,\ (v_c, v_i)\in E_t$ from  the corresponding enhanced node features. We then pass the current node features and enhanced node features to an attention layer, where $h^q=h^c, h^{k,v}=\hat{h}^n$, concatenate its output with $h^c$, and project it back to a $d$-dimensional feature vector. We term this vector the \textit{enhanced current node features} $\hat{h}^c$. After that, we pass the enhanced current node features and neighboring features to a pointer layer~\cite{vinyals2015pointer}, an attention layer directly outputting the attention weights $w$ as the output with $h^q=\hat{h^c},\ h^{k,v}=h^{nb}$. We finally take the output of this pointer layer as the robot's policy, i.e., $\pi_{\theta}(a_t \mid o_t)=w_{i}$.

\end{description}

%%%%%%%%%%%%%%%%%%%%%%%%%%%%%%%%%%%%%%%%%%%%%%%%%%%%%%%%%%%%%%%%%%%%%%%%%%%%%%%%

\subsection{Critic Network}

We train the policy network using the soft actor critic (SAC) algorithm~\cite{christodoulou2019soft,haarnoja2018soft} (see details below), where a critic network is trained to predict state-action values. Since state-action values approximate long-term \textit{returns} (the accumulated sum of rewards), we believe that they also implicitly predict potential gains (i.e., potential areas that might be found), which further helps the robot sequence non-myopic decisions. In practice, we train a critic network to approximate soft state-action values $Q_\phi(o_t, a_t)$, where $\phi$ denotes the set of weights of the critic network. The structure of the critic network is nearly the same as the policy network, except that there is no pointer layer at the end of the decoder. Instead, we directly concatenate the enhanced current node features and neighboring features, then project them to soft state-action values.

%%%%%%%%%%%%%%%%%%%%%%%%%%%%%%%%%%%%%%%%%%%%%%%%%%%%%%%%%%%%%%%%%%%%%%%%%%%%%%%%

\subsection{Training}

\begin{description}[leftmargin=*]

\item[Soft Actor-critic]
SAC aims to learn a policy that maximizes \textit{return} while keeping its entropy as high as possible:
\begin{equation}
    \pi^*= \mathop{\rm {argmax}} \mathbb{E}_{(o_t, a_t)}[\sum^T_{t=0}\gamma^t(r_t+\alpha \mathcal{H}(\pi(.|o_t)))],
\end{equation}
where $\pi^*$ is the optimal policy, $T$ the number of decision steps, $\gamma$ the discount factor, and $\alpha$ the temperature parameter that tunes the importance of the entropy term versus the return.
In SAC, the soft state value is calculated as:
$V(o_t) = \mathbb{E}_{a_t}[Q(o_t,a_t)]-\alpha {\rm log}(\pi(a_t|o_t))$.

The critic loss is calculated as:
$J_Q(\phi)=\mathbb{E}_{o_t}[\frac{1}{2}(Q_\phi(o_t,a_t)-(r_t+\gamma \mathbb{E}_{o_{t+1}}[V(o_{t+1})]))^2]$.

The policy loss loss is calculated as:
$J_\pi(\theta)=\mathbb{E}_{(o_t,a_t)}[\alpha {\rm log}(\pi_\theta(a_t|o_t))-Q_\phi(o_t,a_t)]$.

The temperature parameter is auto-tuned during the training and the temperature loss is calculated as:
$J(\alpha)=\mathbb{E}_{a_t}[-\alpha({\rm log}\pi_t(a_t|o_t)+\overline{\mathcal{H}})]$,

where $\overline{\mathcal{H}}$ denotes the target entropy. In practice, we use double target networks for the critic network training, as in~\cite{christodoulou2019soft,haarnoja2018soft}.

\begin{table*}[tb]
\caption{\textbf{Comparison with baseline ARE planners (100 scenarios for each test set).} We report the average and standard deviation of the trajectory length to complete exploration (lower is better). For utility-based methods~\cite{gonzalez2002navigation}, the numbers 1, 10, 25 represent the value of $\lambda$, which is used to tune exploitation and exploration.}
\vspace{-0.2cm}
\label{table:1}
\begin{center}
\begin{tabular}{c|c|c|c|c|c|c|c|c}
\toprule
 & Nearest & Utility 1 & Utility 10 & Utility 25  & NBVP & TARE Local & CNN & ARiADNE\\
\toprule
easy & 772($\pm$253) & 736($\pm$266) & 732($\pm$256) & 764($\pm$258) & 745($\pm$268) & 692($\pm$228) & 779($\pm$281) & \textbf{663}($\pm$257)\\
medium & 1248($\pm$295) & 1266($\pm$311) & 1179($\pm$300) & 1227($\pm$307)  & 1217($\pm$271) & 1170($\pm$275) & 1169($\pm$319) & \textbf{1130}($\pm$334)\\
complex & 1669($\pm$332) & 1873($\pm$457) & 1662($\pm$347) & 1711($\pm$352) & 1744($\pm$366) & 1646($\pm$312) & 1647($\pm$422) & \textbf{1599}($\pm$363) \\
random & 1354($\pm$410) & 1423($\pm$466) & 1268($\pm$396) & 1315($\pm$413) & 1323($\pm$371) & 1266($\pm$388) & 1323($\pm$428) & \textbf{1204}($\pm$378) \\
\bottomrule
\end{tabular}
\end{center}
\vspace{-0.9cm}
\end{table*}

\item[Training Details]

We utilize the same environments provided in~\cite{chen2019self} for training, which are generated by a random dungeon generator. Each environment is a $640\times480$ grid map, while the sensor range $d_s=80$. To build the collision-free graph, 900 points are uniformly distributed to cover the whole environment, with all points in the known free area considered as candidate viewpoints $V$. We check the $k=20$ nearest neighbor of each viewpoint, and connect them if such an edge is collision-free, to form the edge set $E$. We consider the exploration task to be completed once more than 99\% of the ground truth has been explored ($|P_k|/|P_g| > 0.99$).
During training, we set the max episode length to $128$ decision steps, the discount factor to $\gamma=1$, the batch size to $256$, and the episode buffer size to $10,000$. Training starts after the episode buffer collects more than $2000$ steps data. The target entropy is set to $0.01\cdot{\rm log}(k)$. Each training step contains $1$ iteration and happens after $1$ episode finishes. We use the Adam optimizer with a learning rate of $10^{-5}$ for both policy and critic networks. The target critic network updates every $256$ training steps.
Our model is trained on a workstation equipped with a i9-10980XE CPU and an NVIDIA GeForce RTX 3090 GPU. We train our model utilizing Ray, a distributed framework for machine learning~\cite{moritz2018ray}, to parallelize and accelerate data collection (32 instances in practice). The training needs around 24h to converge. We will release our full code upon acceptance.

\end{description}

%%%%%%%%%%%%%%%%%%%%%%%%%%%%%%%%%%%%%%%%%%%%%%%%%%%%%%%%%%%%%%%%%%%%%%%%%%%%%%%%
%%%%%%%%%%%%%%%%%%%%%%%%%%%%%%%%%%%%%%%%%%%%%%%%%%%%%%%%%%%%%%%%%%%%%%%%%%%%%%%%

\section{EXPERIMENT}

\begin{figure}[tb]
    \centering
    \subfloat[simple]{\includegraphics[width=0.15\textwidth]{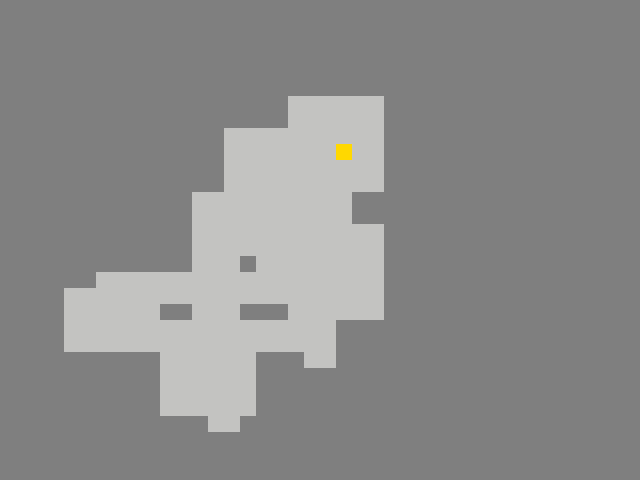}}
    \hfill
    \subfloat[medium]{\includegraphics[width=0.15\textwidth]{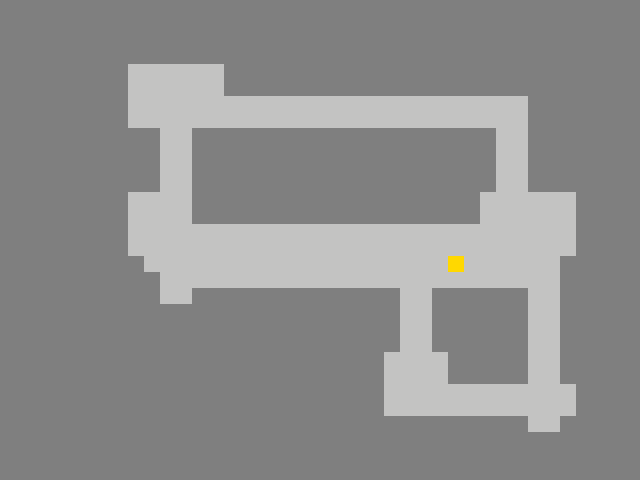}}
    \hfill
    \subfloat[complex]{\includegraphics[width=0.15\textwidth]{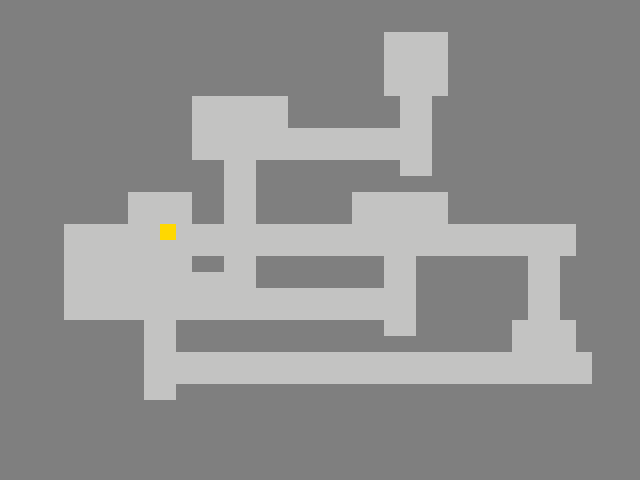}}
    \vspace{-0.2cm}
    \caption{\textbf{Examples scenarios from each different test set.}}
    \label{fig:test_set_examples}
    \vspace{-0.4cm}
\end{figure}

%%%%%%%%%%%%%%%%%%%%%%%%%%%%%%%%%%%%%%%%%%%%%%%%%%%%%%%%%%%%%%%%%%%%%%%%%%%%%%%%

\subsection{Comparison Analysis}

Most previous works often only conduct experiments in a few scenarios (often less than 10). However, we note that the performance of exploration planners exhibits high variance in different scenarios. Therefore, we believe a convincing comparison should be based on evaluation in a large number of testing environments. Although building so many testing environments is tricky and time-consuming even in ROS, hundreds of simplified scenarios, like the ones we used for training, can be generated easily. Therefore, we conduct comparison analyses on a fixed set of simplified environments, which were never seen by our trained model.
Testing environments are divided in four sets (100 scenarios each), named \textit{random}, \textit{easy}, \textit{medium}, and \textit{complex}. \textit{Easy} scenarios only contain one room, and \textit{complex} scenarios contain multiple rooms with complicated corridors, while the complexity of \textit{medium} scenarios lies in-between. \textit{Random} scenarios contain a mix of \textit{easy}, \textit{medium}, and \textit{complex} scenarios (but no repeated scenario from these test sets).

\begin{figure}[tb]
    \vspace{0.2cm}
    \centering
    \subfloat[Trajectory Analysis]{\includegraphics[width=0.26\textwidth]{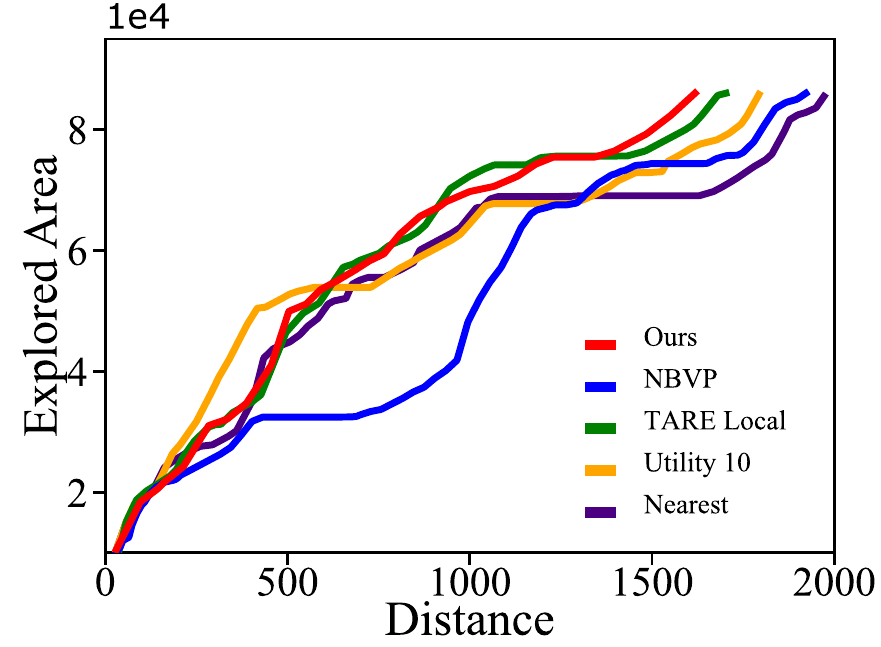}}
    \hfill
    \subfloat[ARiADNE (1618)]{\includegraphics[width=0.22\textwidth]{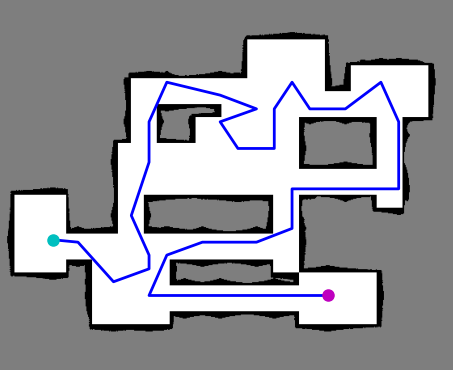}}
    \vfill
    \subfloat[TARE Local (1703)]{\includegraphics[width=0.15\textwidth]{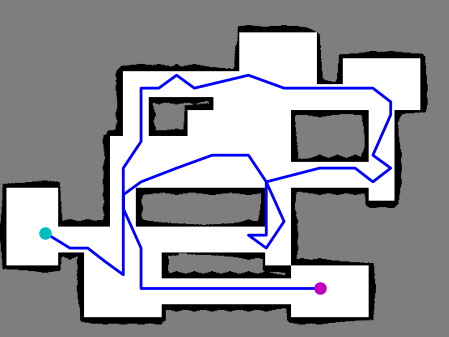}}
    \hfill
    \subfloat[NBVP (1922)]{\includegraphics[width=0.15\textwidth]{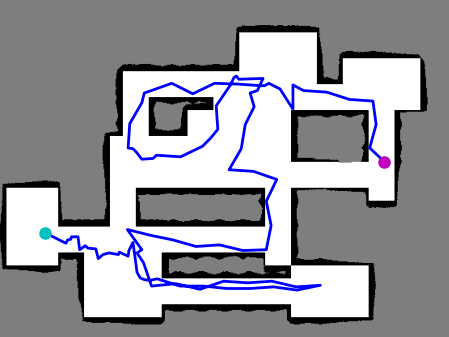}}
    \hfill
    \subfloat[Utility 10 (1793)]{\includegraphics[width=0.15\textwidth]{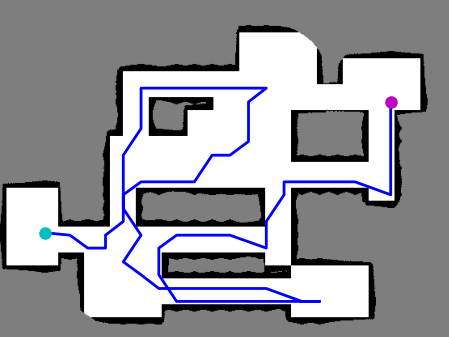}}
    \hfill
    \vspace{-0.2cm}
    \caption{\textbf{Visual comparison of our method and baselines in an example scenario.}}
    \label{fig:comparison}
    \vspace{-0.4cm}
\end{figure}

We compare ARiADNE with state-of-the-art conventional planners, including Nearest Frontier~\cite{yamauchi1997frontier}, Utility-based Frontier~\cite{gonzalez2002navigation}, NBVP~\cite{bircher2016receding}, and TARE Local~\cite{cao2021tare}. Nearest Frontier always drives the robot towards the nearest frontier, while Utility-based Frontier evaluates the gain of each frontier $g_i=u_i \cdot e^{-\lambda \cdot {\rm C}(\psi_i)}$ and drives the robot to the frontier with the highest gain, where $u_i$ is the utility of frontier $i$, $\psi_i$ the shortest path from the robot's current position to frontier $i$, and $\lambda$ a tunable parameter used to balance exploration and exploitation. The same function is also used in NBVP to evaluate sampled trajectories. We tried a series of values of $\lambda$ for Utility-based Frontier and NBVP, and found that $\lambda=10$ generally performs best (see Table~\ref{table:1}). Finally, TARE Local refers to the local planner of TARE~\cite{cao2021tare}, which explicitly plans a full trajectory to cover all frontiers (we do not use TARE's global planner, since its local planning horizon already fits our testing environments). NBVP and TARE run 300 and 10 iterations for each decision step respectively (15 and 1 in default~\cite{bircher2016receding,cao2021tare}), to make their decisions as optimal as possible. In our tests, we adopt our collision-free graph as the trajectory space for all baselines except NBVP (we found RRTs mostly generate poor zig-zag paths due to symmetries in our uniform graph), to alleviate the randomness of sampling and ensure a fair comparison. We further compare against a CNN-based DRL planner~\cite{chen2019self}. Since this CNN-based planner has a fixed observation range, it only has a partial observation of the (partial) map, and relies on a frontier-based method for exploration when there is no nearby frontier in its field-of-view.

We report the average and variance of the total trajectory length to complete exploration in Table~\ref{table:1}. Our results indicate that ARiADNE outperforms all baselines on average, in all test sets. We do not report the planning time of baseline methods in Table~\ref{table:1}, since we focused on implementing fundamental inner workings of the baselines, without perfectly optimizing their computing time. In addition, we observed that the utility/gain computation generally takes 90\% of the planning time for conventional methods in practice, while its computing time is determined by the resolution of the map. Therefore, computing times vary greatly in different exploration scenarios. Despite this, we note that our method can be used in real-time. Under our exploration setting, our method takes 0.7s for the observation generation on average (utility computation and graph building) and less than 0.02s for the neural network inference on a i9-10980XE CPU.

\begin{figure}
    \centering
    \includegraphics[width=0.36\textwidth]{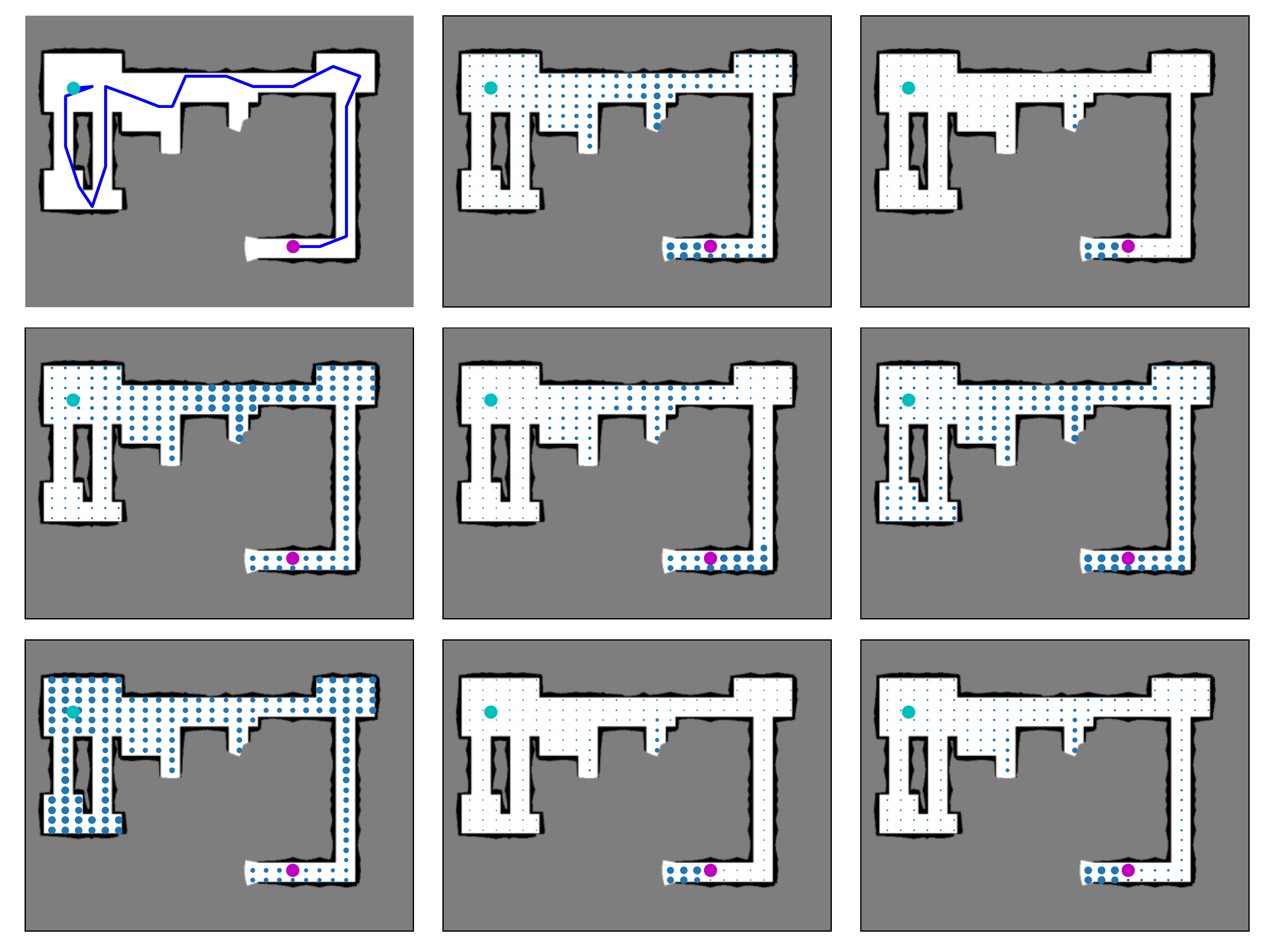}
    \vspace{-0.2cm}
    \caption{\textbf{Attention weights visualization of the critic network decoder.} The query vector is the node at the current robot's position (purple) and the keys vector are nodes in the augmented graph (blue). Note how the different heads of the decoder learn to focus on either local or global dependencies of areas in the partial map.}
    \label{fig:attention_weights}
    \vspace{-0.45cm}
\end{figure}

As discussed in the related work section, the best-tuned frontier-based method (Utility 10) performs well in 2D exploration tasks (better than NBVP). Despite this, since these frontier-based methods are myopic, they are outperformed by TARE Local, which plans near-optimal long-term (full) trajectories on the current partial map. While it only constructs paths one viewpoint at a time, our learning-based method can not only reason about the whole partial map to construct efficient, non-myopic exploration trajectories, while learning to predict the potential long-term gain of decisions. We believe such an advantage results in the improvement of our method over conventional baselines (5\% better than TARE Local in our \textit{random} scenarios). 
Fig.~\ref{fig:comparison} shows an example where ARiADNE plans a more efficient trajectory, while conventional methods suffer from redundant movements. However, it should be noted that considering long-term paths and predicting potential gains do not strictly guarantee better performance in every scenario (e.g., predictions could be wrong). In fact, ARiADNE plans the shortest path for 33 scenarios in our \textit{random} tests, while TARE Local, NBVP, Utility 10 perform best in 23, 21, 23 scenarios respectively.   

Finally, ARiADNE also outperforms the CNN-based planner. We believe that our main advantage stems from the attention-based neural network, which efficiently learns features at different scales (as shown in Fig.~\ref{fig:attention_weights}, different heads of the decoder learn to focus on either local or global dependencies), while CNNs naturally only focus on local dependencies. Therefore, our model can better learn dependencies between different areas to reason about the entire partial map and avoid myopic decisions.  

\begin{figure}[tb]
\vspace{-0.3cm}
    \centering
    \subfloat[Ground truth]{\includegraphics[width=0.23\textwidth]{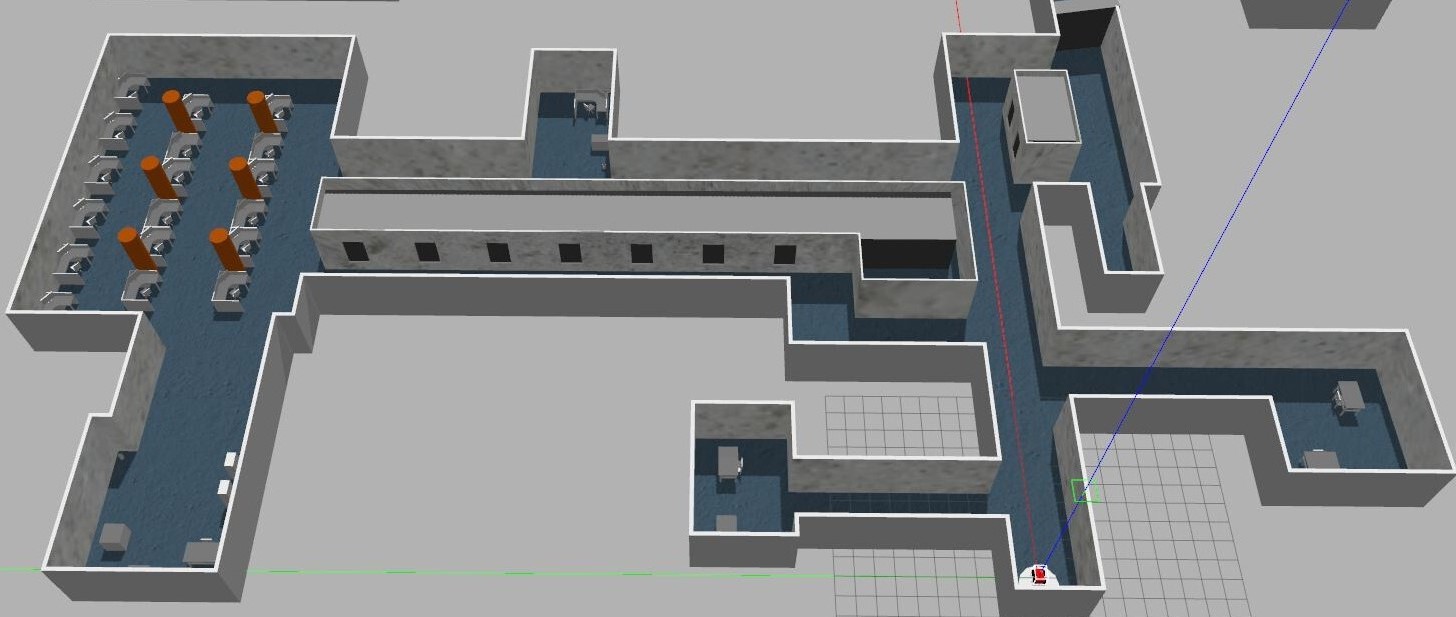}}
    \hfill
    \subfloat[Constructed Octomap]{\includegraphics[width=0.25\textwidth]{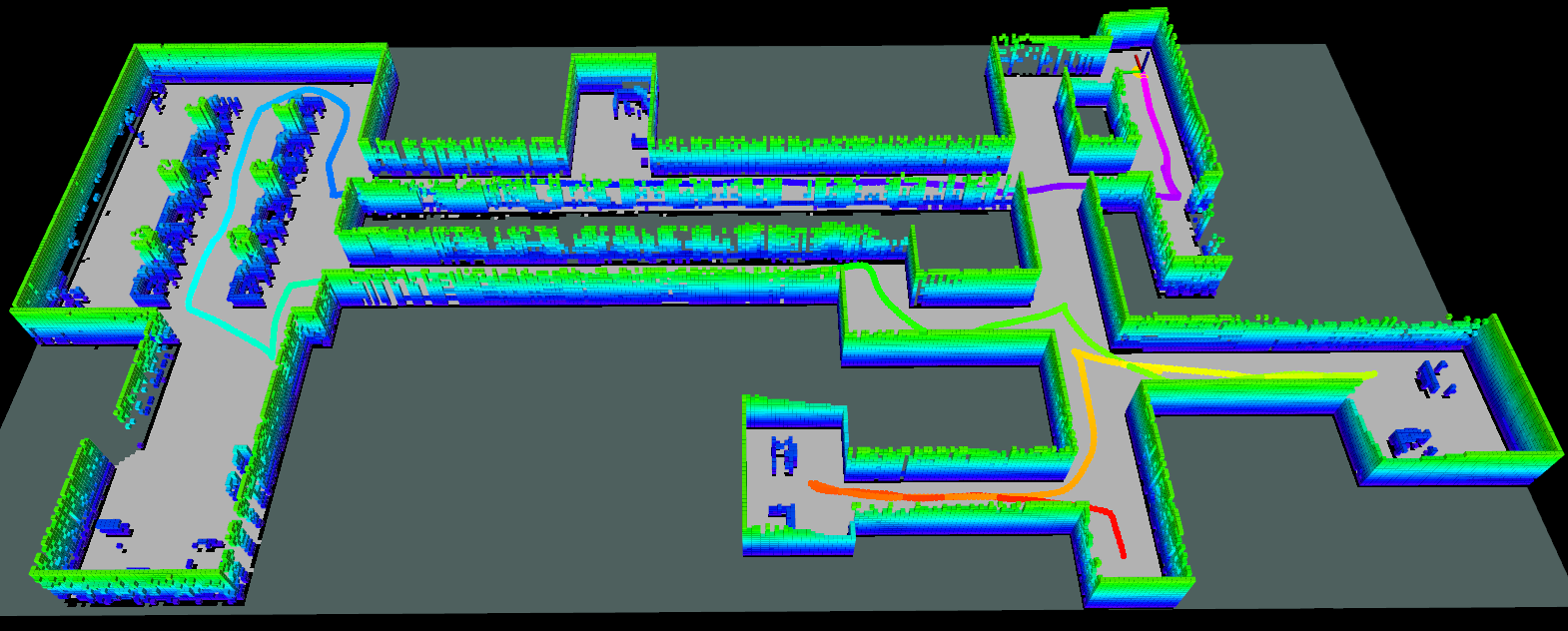}}
    \vfill
    \vspace{-0.3cm}
    \subfloat[Constructed occupancy grid map]{\includegraphics[width=0.30\textwidth]{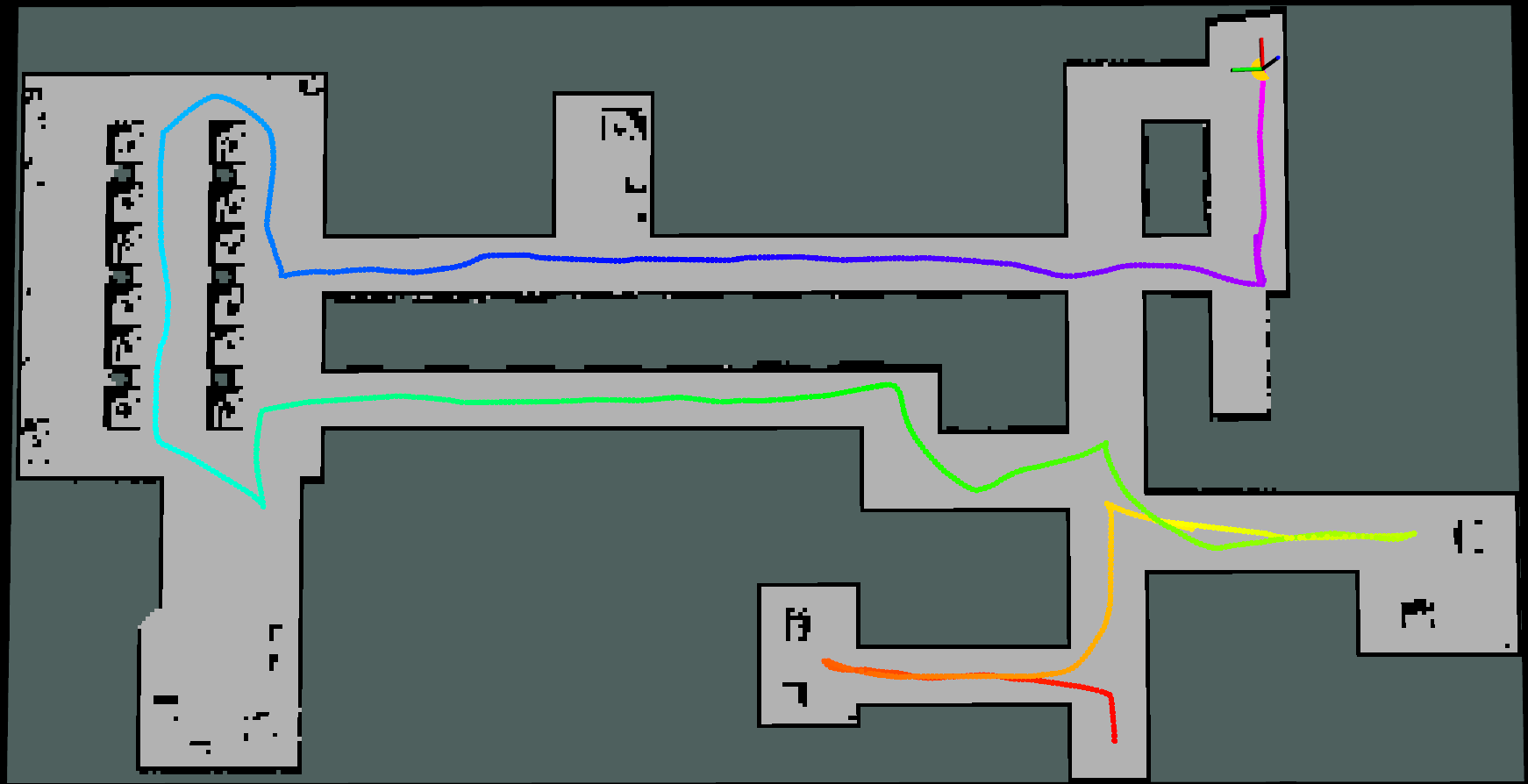}}
    \vspace{-0.2cm}
    \caption{\textbf{Validation of our method in simulation.} Note that ignoring the small left-down corner is actually a wise decision since the objective is to explore 99\% of the environment.}
    \label{fig:validation}
    \vspace{-0.45cm}
\end{figure}

%%%%%%%%%%%%%%%%%%%%%%%%%%%%%%%%%%%%%%%%%%%%%%%%%%%%%%%%%%%%%%%%%%%%%%%%%%%%%%%%

\subsection{Experimental validation}

We validate ARiADNE in a simulation environment for exploration provided by~\cite{cao2022autonomous}. It contains fundamental modules (e.g., state estimation and motion control), which allow us to consider a real sensor model and a low-level motion controller. The validation is conducted in a realistic indoor environment (approximately 70m $\times$ 40m) with long and narrow corridors connected with tables, colums, and lobby areas (see Fig.~\ref{fig:validation}(a)).
We use a wheeled robot equipped with a 3D Velodyne Lidar with a 130m sensor range. We convert collected data into an Octomap (see Fig.~\ref{fig:validation}(b)) and then project it to a occupancy grid map for exploration planning (see Fig.~\ref{fig:validation}(c)). The resolution of the grid map is 0.2m. We re-plan the path every 0.2s. 
Although the sensor model (i.e., sensor range and sensing frequency) of the robot is drastically different from the one used in training, our trained model still makes efficient decisions to avoid redundant movements for exploration (see the colored trajectory in Fig.~\ref{fig:validation}(c), highlighting the generalizability of our approach.

%%%%%%%%%%%%%%%%%%%%%%%%%%%%%%%%%%%%%%%%%%%%%%%%%%%%%%%%%%%%%%%%%%%%%%%%%%%%%%%%
%%%%%%%%%%%%%%%%%%%%%%%%%%%%%%%%%%%%%%%%%%%%%%%%%%%%%%%%%%%%%%%%%%%%%%%%%%%%%%%%

\section{CONCLUSION}

In this work, we propose ARiADNE, a reinforcement learning approach that relies on attention-based deep neural network for autonomous exploration. Our approach allows the robot to efficiently learn dependencies between different areas in its partial map and implicitly predict potential gains, thus allowing it to sequence non-myopic movement decisions in partially-known environments. In our tests, ARiADNE exhibits improvement over state-of-the-art frontier-based, sampling-based, and CNN-based exploration planners, in terms of average trajectory length to complete exploration. We also validate our approach in a high-fidelity ROS simulation, where we consider a real sensor model and a low-level motion controller, towards deployments on real robots.

Future work will focus on extending our approach to autonomous exploration of 3D environments, where frontiers are much denser than in 2D. Second, although in this work we uniformly distribute nodes to construct a graph, we believe a sparser graph containing more informative viewpoints may improve performance. Finally, we are also interested in explicitly predicting the potential gain during exploration to further boost planning performance.

%%%%%%%%%%%%%%%%%%%%%%%%%%%%%%%%%%%%%%%%%%%%%%%%%%%%%%%%%%%%%%%%%%%%%%%%%%%%%%%%
%%%%%%%%%%%%%%%%%%%%%%%%%%%%%%%%%%%%%%%%%%%%%%%%%%%%%%%%%%%%%%%%%%%%%%%%%%%%%%%%

\section*{Acknowledgments}

This work was supported by Temasek Laboratories (TL@NUS) under grant TL/FS/2022/01.

\bibliographystyle{IEEEtran}
\bibliography{ref.bib}

\begin{thebibliography}{10}
\providecommand{\url}[1]{#1}
\csname url@rmstyle\endcsname
\providecommand{\newblock}{\relax}
\providecommand{\bibinfo}[2]{#2}
\providecommand\BIBentrySTDinterwordspacing{\spaceskip=0pt\relax}
\providecommand\BIBentryALTinterwordstretchfactor{4}
\providecommand\BIBentryALTinterwordspacing{\spaceskip=\fontdimen2\font plus
\BIBentryALTinterwordstretchfactor\fontdimen3\font minus
  \fontdimen4\font\relax}
\providecommand\BIBforeignlanguage[2]{{%
\expandafter\ifx\csname l@#1\endcsname\relax
\typeout{** WARNING: IEEEtran.bst: No hyphenation pattern has been}%
\typeout{** loaded for the language `#1'. Using the pattern for}%
\typeout{** the default language instead.}%
\else
\language=\csname l@#1\endcsname
\fi
#2}}

\bibitem{hornung13auro}
\BIBentryALTinterwordspacing
A.~Hornung, K.~M. Wurm, M.~Bennewitz, C.~Stachniss, and W.~Burgard,
  ``{OctoMap}: An efficient probabilistic {3D} mapping framework based on
  octrees,'' \emph{Autonomous Robots}, 2013, software available at
  \url{https://octomap.github.io}. [Online]. Available:
  \url{https://octomap.github.io}
\BIBentrySTDinterwordspacing

\bibitem{placed2022survey}
J.~A. Placed, J.~Strader, H.~Carrillo, N.~Atanasov, V.~Indelman, L.~Carlone,
  and J.~A. Castellanos, ``A survey on active simultaneous localization and
  mapping: State of the art and new frontiers,'' \emph{arXiv preprint
  arXiv:2207.00254}, 2022.

\bibitem{bircher2016receding}
A.~Bircher, M.~Kamel, K.~Alexis, H.~Oleynikova, and R.~Siegwart, ``Receding
  horizon" next-best-view" planner for 3d exploration,'' in \emph{2016 IEEE
  international conference on robotics and automation (ICRA)}.\hskip 1em plus
  0.5em minus 0.4em\relax IEEE, 2016, pp. 1462--1468.

\bibitem{cao2021tare}
C.~Cao, H.~Zhu, H.~Choset, and J.~Zhang, ``Tare: A hierarchical framework for
  efficiently exploring complex 3d environments.'' in \emph{Robotics: Science
  and Systems}, 2021.

\bibitem{yamauchi1997frontier}
B.~Yamauchi, ``A frontier-based approach for autonomous exploration,'' in
  \emph{Proceedings 1997 IEEE International Symposium on Computational
  Intelligence in Robotics and Automation CIRA'97.'Towards New Computational
  Principles for Robotics and Automation'}.\hskip 1em plus 0.5em minus
  0.4em\relax IEEE, 1997, pp. 146--151.

\bibitem{gonzalez2002navigation}
H.~H. Gonz{\'a}lez-Banos and J.-C. Latombe, ``Navigation strategies for
  exploring indoor environments,'' \emph{The International Journal of Robotics
  Research}, vol.~21, no. 10-11, pp. 829--848, 2002.

\bibitem{selin2019efficient}
M.~Selin, M.~Tiger, D.~Duberg, F.~Heintz, and P.~Jensfelt, ``Efficient
  autonomous exploration planning of large-scale 3-d environments,'' \emph{IEEE
  Robotics and Automation Letters}, vol.~4, no.~2, pp. 1699--1706, 2019.

\bibitem{holz2010evaluating}
D.~Holz, N.~Basilico, F.~Amigoni, and S.~Behnke, ``Evaluating the efficiency of
  frontier-based exploration strategies,'' in \emph{ISR 2010 (41st
  International Symposium on Robotics) and ROBOTIK 2010 (6th German Conference
  on Robotics)}.\hskip 1em plus 0.5em minus 0.4em\relax VDE, 2010, pp. 1--8.

\bibitem{kulich2011distance}
M.~Kulich, J.~Faigl, and L.~P{\v{r}}eu{\v{c}}il, ``On distance utility in the
  exploration task,'' in \emph{2011 IEEE International Conference on Robotics
  and Automation}.\hskip 1em plus 0.5em minus 0.4em\relax IEEE, 2011, pp.
  4455--4460.

\bibitem{dang2020graph}
T.~Dang, M.~Tranzatto, S.~Khattak, F.~Mascarich, K.~Alexis, and M.~Hutter,
  ``Graph-based subterranean exploration path planning using aerial and legged
  robots,'' \emph{Journal of Field Robotics}, vol.~37, no.~8, pp. 1363--1388,
  2020.

\bibitem{xu2021autonomous}
Z.~Xu, D.~Deng, and K.~Shimada, ``Autonomous uav exploration of dynamic
  environments via incremental sampling and probabilistic roadmap,'' \emph{IEEE
  Robotics and Automation Letters}, vol.~6, no.~2, pp. 2729--2736, 2021.

\bibitem{arora2017randomized}
S.~Arora and S.~Scherer, ``Randomized algorithm for informative path planning
  with budget constraints,'' in \emph{2017 IEEE International Conference on
  Robotics and Automation (ICRA)}.\hskip 1em plus 0.5em minus 0.4em\relax IEEE,
  2017, pp. 4997--5004.

\bibitem{hollinger2014sampling}
G.~A. Hollinger and G.~S. Sukhatme, ``Sampling-based robotic information
  gathering algorithms,'' \emph{The International Journal of Robotics
  Research}, vol.~33, no.~9, pp. 1271--1287, 2014.

\bibitem{niroui2019deep}
F.~Niroui, K.~Zhang, Z.~Kashino, and G.~Nejat, ``Deep reinforcement learning
  robot for search and rescue applications: Exploration in unknown cluttered
  environments,'' \emph{IEEE Robotics and Automation Letters}, vol.~4, no.~2,
  pp. 610--617, 2019.

\bibitem{schmid2022fast}
L.~Schmid, C.~Ni, Y.~Zhong, R.~Siegwart, and O.~Andersson, ``Fast and
  compute-efficient sampling-based local exploration planning via distribution
  learning,'' \emph{arXiv preprint arXiv:2202.13715}, 2022.

\bibitem{zhu2018deep}
D.~Zhu, T.~Li, D.~Ho, C.~Wang, and M.~Q.-H. Meng, ``Deep reinforcement learning
  supervised autonomous exploration in office environments,'' in \emph{2018
  IEEE international conference on robotics and automation (ICRA)}.\hskip 1em
  plus 0.5em minus 0.4em\relax IEEE, 2018, pp. 7548--7555.

\bibitem{li2019deep}
H.~Li, Q.~Zhang, and D.~Zhao, ``Deep reinforcement learning-based automatic
  exploration for navigation in unknown environment,'' \emph{IEEE transactions
  on neural networks and learning systems}, vol.~31, no.~6, pp. 2064--2076,
  2019.

\bibitem{chen2019self}
F.~Chen, S.~Bai, T.~Shan, and B.~Englot, ``Self-learning exploration and
  mapping for mobile robots via deep reinforcement learning,'' in \emph{Aiaa
  scitech 2019 forum}, 2019, p. 0396.

\bibitem{chen2020autonomous}
F.~Chen, J.~D. Martin, Y.~Huang, J.~Wang, and B.~Englot, ``Autonomous
  exploration under uncertainty via deep reinforcement learning on graphs,'' in
  \emph{2020 IEEE/RSJ International Conference on Intelligent Robots and
  Systems (IROS)}.\hskip 1em plus 0.5em minus 0.4em\relax IEEE, 2020, pp.
  6140--6147.

\bibitem{cao2022catnipp}
\BIBentryALTinterwordspacing
Y.~Cao, Y.~Wang, A.~Vashisth, H.~Fan, and G.~A. Sartoretti, ``{CA}t{NIPP}:
  {C}ontext-{A}ware {A}ttention-based {N}etwork for {I}nformative {P}ath
  {P}lanning,'' in \emph{\textbf{Accepted to} the 6th Annual Conference on
  Robot Learning}, 2022. [Online]. Available:
  \url{https://openreview.net/forum?id=cAIIbdNAeNa}
\BIBentrySTDinterwordspacing

\bibitem{vaswani2017attention}
A.~Vaswani, N.~Shazeer, N.~Parmar, J.~Uszkoreit, L.~Jones, A.~N. Gomez,
  {\L}.~Kaiser, and I.~Polosukhin, ``Attention is all you need,''
  \emph{Advances in neural information processing systems}, vol.~30, 2017.

\bibitem{vinyals2015pointer}
O.~Vinyals, M.~Fortunato, and N.~Jaitly, ``Pointer networks,'' \emph{Advances
  in neural information processing systems}, vol.~28, 2015.

\bibitem{dwivedi2020generalization}
V.~P. Dwivedi and X.~Bresson, ``A generalization of transformer networks to
  graphs,'' \emph{arXiv preprint arXiv:2012.09699}, 2020.

\bibitem{christodoulou2019soft}
P.~Christodoulou, ``Soft actor-critic for discrete action settings,''
  \emph{arXiv preprint arXiv:1910.07207}, 2019.

\bibitem{haarnoja2018soft}
T.~Haarnoja, A.~Zhou, K.~Hartikainen, G.~Tucker, S.~Ha, J.~Tan, V.~Kumar,
  H.~Zhu, A.~Gupta, P.~Abbeel, \emph{et~al.}, ``Soft actor-critic algorithms
  and applications,'' \emph{arXiv preprint arXiv:1812.05905}, 2018.

\bibitem{moritz2018ray}
P.~Moritz, R.~Nishihara, S.~Wang, A.~Tumanov, R.~Liaw, E.~Liang, M.~Elibol,
  Z.~Yang, W.~Paul, M.~I. Jordan, \emph{et~al.}, ``Ray: A distributed framework
  for emerging ai applications,'' in \emph{Proceedings of OSDI}, 2018, pp.
  561--577.

\bibitem{cao2022autonomous}
C.~Cao, H.~Zhu, F.~Yang, Y.~Xia, H.~Choset, J.~Oh, and J.~Zhang, ``Autonomous
  exploration development environment and the planning algorithms,'' in
  \emph{2022 International Conference on Robotics and Automation (ICRA)}.\hskip
  1em plus 0.5em minus 0.4em\relax IEEE, 2022, pp. 8921--8928.

\end{thebibliography}

\end{document}